***Between collective intelligence and semantic web : hypermediating sites. Contribution to technologies of intelligence***

Lise VERLAET

Associate professor in Information and Communication Sciences,

LERASS-Céric, Paul-Valéry University – France

Mail : lise.verlaet@univ-montp3.fr

Phone : +33 (0)651 06 07 78

Sidonie GALLOT

Associate professor in Information and Communication Sciences,

LERASS-Céric, Paul-Valéry University – France

Mail : sidonie.gallot@univ-montp3.fr

Phone : +33 (0)699 02 77 04

LERASS-Céric (EA 827)

Université Paul-Valéry

Route de Mende

34 199 Montpellier Cedex 5

France

***Between collective intelligence and semantic web : hypermediating sites. Contribution to technologies of intelligence***

*Abstract* — In this paper we present a new form of access to knowledge through what we call « hypermediator websites ». These hypermediator sites are intermediate between information devices that just scan the book culture and a "real" hypertext writing format.

*Keywords* – Hypermediator website, semantic web, collective intelligence, model markup systems, Information Visualization, technology of intelligence.

**1. Introduction**

The two last decades have propelled ours societies in the digital age and his contemporaries in the hyperconnexion. The advent of information technology and digital communication seemed like a great opportunity in terms of diffusion and transmission of knowledge. Were we not supposed to live a knowledge revolution? The Web pioneers would finally see their dream come true : increase human cognitive abilities, allow the humanity to increase its knowledge and thus accelerate its evolution?

If many efforts are made in this direction, we must admit that the dissemination procedures adopted by new technologies are, for the most part, far removed from issues of knowledge transmission. Simple digitalization for some, always in the grips of the bookish volume (culture of which it is difficult to emancipate after several millenniums of reign). For others, it is the deviances of simplified and standardized uses of technical tools which neglect partially their potential to serve the knowledge. Finally, let us not forget that the tools of the social Web are staying resolutely based on telecommunication principles.

If the technique can theoretically do, nowadays, almost everything, it is to think the dynamic organization and management of knowledge by querying such "natural" models of knowledge construction process in order to think it in adequacy the way to organize it "artificially". The anthropologists of the communication demonstrated that the linguistic thought is interdependent of the anthropological thought, itself correlative to the ontological thought. The semantic of a knowledge domain constitute, by essence, the musical notes used in the composition of the harmony of a culture, following the example of the orchestral metaphor (Winkin, 1981, p 5-6). This synchronization is the reflection of the culture, which is dependent on the environment within which evolve the actors, dependent on their ecosystem. As indicates Goody (2007) the language is a "technology of the intellect", in this perspective, semantics, understood as the study of a language and its system of meaning, is at the heart of this technology of the intellect, this one that has allowed the humanity to evolve and to get organized by the information and the communication.

In this sense, our reflection rejoins the formal approaches of the hypertext (Rety, 2005) to develop new forms of technology intelligence through the concept of "hypermediator website", which is based on the principles of collective intelligence (Levy, 1994) and operates according to the precepts and semantic Web tools. After reviewing the theoretical and practical models that brought us to conceive this hypermediator website, we will explain the possibilities of searching and browsing proposed by this device.

## 2. Theorical and practical models of « hypermediating sites »

### *2.1 Formal approaches of hypertext*

With the rise in power of the Internet and the democratization of its tools, hypertext has become synonymous with "hyperlink". If the latter is omnipresent on the Web and in digital devices, its use is far away from that imagined by the precursors of the Web when they invented the hypertext concept, its role is now often summarized to an interactive table of contents or a bookmark. Following the example of Rety (2005), we believe that a return to formal approaches of the hypertext is necessary in order to reposition scientifically this concept in its early definitions. Hypertext is a way to forge constructive relations between ideas, between units of sense stemming from a non-sequential writing. It has to be considered as a set of fragments of texts of variable granularity and semantically interconnected. In fact, the "hyperlink" should be thought of as a part of the hypertext. This non-sequentiality is described as both the most likely to match with the cognitive mnemonic associations human system but also as the foundation of its own digital writing (Bush, 1945, Nelson 1981, 1999).

In the lineage of the pioneers, we see in the interface "man / machine" a considerable opportunity to increase, to transmit and manage knowledge. We support the idea that the hypertext can account the complex reality of a knowledge area, that it allows to accompany human reasoning.

*2.2 Principles of Semantic Web*

Aware of the excesses of the current Web, Berners-Lee and the W3C launched in early 2000 the project of the Semantic Web. "*The Semantic Web is an extension of the current web in which information is given well-defined meaning, better enabling computers and people to work in cooperation*" (Berners-Lee & al., 2001). The aim of the latter is simple, it is to give to machines a semantic scope by tagging digital contents using XML, and organizing it with RDF and ontology layers of its architecture. This having the effect of clarifying, organizing information content, facilitate processing by computer systems and thereafter consultation by users. Another purpose - and not the least - of the Semantic Web project is to ensure interoperability between computer systems for sharing and reuse of information. Knowing that the culmination of the Semantic Web, according to its creators, is home automation, that is to say the set of automated techniques and technologies facilitating the daily human.

We have dealt a particular interest in XML - whose flexibility and manoeuvrability really allow us to think a hypertext system - and the organization of knowledge.

*2.3 Principles of collective intelligence*

Lévy (2002, p.243-245) defines intelligence as a "*power of self-creation*", which develops from a cognitive point of view through the "*capacity for autonomous learning*" and an "*evolution process*". It emerges from "*circular interaction process and self-generator between a large number of complex systems*". In fact, "*the intelligence is always the result of a collective numerous and interdependent*", he underlines the pleonastic character of "collective intelligence". Lévy states that "*major stages of cultural evolution correspond to mutations in the process of collective intelligence, almost always* *linked (in a complex way, and on circular causality mode) with mutations in the life of language*" and insists that "*language is precisely what makes culture - that is to say the collective intelligence deliberately working to its own development – possible*". Lévy also raises problematics inherent to cultural prejudices postulating that intelligence would be "*the property of individuals*", phenomenon that has important resonances in the scientific and technical community (among others).

We also abound in the direction of Lévy when it states that the digital sphere is an area more than favourable to study the knowledge emanating from the collective intelligence and thus to make emerge current trends. Lévy's vision on collective intelligence is also reminiscent of the orchestra model advanced by Winkin (1981), itself inspired by the theatrical metaphor of Goffman (1975): everyone must listen and grant to others in order to form a melody in unison (dominant culture), the dissonances sometimes making part of the game (subsidiary crops).

Our works lean on collective intelligence to uncover governance of knowledge that governs a given domain.

*2.4 Principles of "hypermediator websites"*

Our purpose envisages the governance of knowledge as an "artifact" of human thought system, as an extension of the memory. The process of transmission of knowledges, the knowledges governance appears as the result of a mediation between the "reader", his system of organizing knowledge, the "learned society" given system of organizing knowledge, in which the technical tool is only a mediator to facilitate the process, tool conceived and organized by men and for men aiming to facilitate access to knowledge. Through this governance of knowledge, we seek to develop new forms of intelligence technologies.

The principles of the hypertext, of the Semantic Web and of the collective intelligence form a theoretical and practical background on which we have nourished ourselves to develop the idea of "hypermediator websites". We define a "hypermediator websites" as an artefact space of mediation allowing linking: a field of knowledge, non-sequential knowledges, a reader, a mediator-tagger. It is found to be an intermediate device between a digitalization of the book culture (the current Web) and a “real” writing hypertext (web semantically organized of tomorrow). An intermediate device because it is not intended - at least for now - to be a standalone device, but should be more considered as a “mediator” website (Davallon & Jeanneret, 2004). However, an important distinction is to be made between “mediator website” and “hypermediator website”. If the first one officiates as a portal website composed of hyperlinks that point to other external portal websites, the second is conceived as a complementary and intrinsic website to a collection of documents that offering an effective treatment of his corpus in order to reach a new meaning.

A hypermediator website do not only consider the full texts or the elements of the paratext but also units of meaning on concepts that compose them. As specified by Clément “the layer of the conceptual hypertext which is the most characteristic and that justifies the appellation of “intellectual technology” because by organizing the data in a system that binds them together, it gives it a meaning and product a new information” (2007, p.3). Hypermediator website intends to allow the reader to learn, to compare, to compare ideas, to understand the interactions between concepts (etc.) and thus gives place to a further reflection but also finer of a knowledge domain.

The hypermediator website therefore plans to return, through specific storyboarding, the knowledges stemming from the collective intelligence of the authors represented in the processed digital documentary background. To reach this goal, it is necessary firstly to conduct a semantic markup of the corpus. This step allows us to collect not only data but information on these data. Indeed, the data do not really interest taken in isolation, it is only once collected, organized and interrelated that it take all its sense, that it become information (Mazza, 2009). As we will see, the markup model that we propose notes and reveals enough contextual clues on the informations so that they become knowledges. The confrontation between the received information and our *relevance system* (Schütz, 1987) or our *framework* (Goffman & Kihm, 1974) ensures the construction of meaning and the development of a contextual intelligence.

The particularity of our work rely on the approaches chosen to apprehend, process and analyze the corpus. Indeed we took the party to adapt analysis tools generally used to study situations and communication systems between social actors to transpose it to their writings and products of their knowledge. Furthermore, it seems to be important to note that we are interested in the scientific literature and in particular scientific papers. We think like Farchy et al. (2010, p.10) that the "*article retains its position of fundamental building block into modern science*".

## 3. Hypermediator website's markup model

The hypermediator website is based on the markup model by the Semio-Contextual Approach of Corpus (SCAC), model formulated from an approach of contextualizing information : the semiotic context analysis (Mucchielli, 2005) which is a method of contextual and cognitive analysis of communication situations. For the purposes of this article and our demonstration, we will only expose a portion of the markup model, the one dedicated to the contents of the corpus, to the units of meaning concerning concepts. To know that the SCAC model is a generic model and it considers the same elements whatever it applies on the paratexts, on the thematics or the authors of the articles collection (Verlaet, 2011). We will not address either the transposition of the semio-contextual analysis of communication situations to the ASCC's markup model (Verlaet, 2010).

### *3.1 ASCC's model markup*

The SCAC's markup model was designed to identify the relevant units or fragments of meaning on concepts and meet the readers information needs in a search related in encyclopedic or terminologic issues of a knowledge field. *"The culture in the anthropological sense can be summarily understood as a reading grid of the world and therefore a matrix of representations, of knowledges, of beliefs and behaviors transmitted by the family, by the group or society"* (Max Weber quoted by Engelhard, 2012, p.135). The contexts highlighted by semio-contextual analysis act as a reading grid which have to make emerge the sens of a communication situation between actors. We have implemented this reading grid within the SCAC markup model to make emerge the meaning of concepts present in scientific articles.

| Tags | Units of meaning |
|---|---|
| Identity | Identify a significant fragment on a concept |
| <norm id="concept X"> | Fragment of informations stating the definition of the concept. |
| <stakes id="concept X"> | Fragment of informations explaining the stakes, the objectives of the concept. |
| <position holonym="concept X" meronym="concept Y"> or <position hypernym="concept X" hyponym="concept Y"> | Fragment of informations marking a specification relation between concepts. |
| <relations a="concept X" b = "concept Y" type = "link"> | Fragment of information exposing relations (except specification). |
| <time id="concept X" date="2009"> | Fragment of information disclosing historical clues of the concept. |
| <spatial id="concept X" lieu="France"> | Fragment of information correlating a place with a concept. |
| <quote id="concept X" auteur="auteur" reference="reference"> | Fragment of information indicating a quotation from an author on a concept. |

*Fig. 1. XML markup grid of concepts according to the SCAC model*

The table above therefore describes the XML markup grid that the tagger have to follow, it being understood that the identification of the meaning can only be revealed by a human action.

*3.2 The mediator-tagger*

The role of the tagger is very important. It is essential that he is a domain expert since it is necessary to fully understand the conceptual universe inherent to the scientific corpus. It is the expertise of the tagger what ensures the relevance of fragments of information. The tagger position is also tactful because his expertise directs his reading, which is necessarily under the influence of his social framework, of his reference framework. All the difficulty lies in the mobilization of his reference framework while respecting the thoughts and writings of the authors. The tagger is considered an actor who has the faculties to understand the corpus but who is also sufficiently neutral and honest to be the mediator between the author and the reader. The tagger is seen as performing a fine architect job to mediate knowledge in a shared common framework that he helps to define and, as we will see, to model.

This is not the tagger which is in charge of evaluation the reliability or validity of scientific contents, his task is only focused on identifying relevant text fragments according to the SCAC grid markup. The reliability and validity of concepts and their interconnections are provided by the collective intelligence of the authors, by its recurrence and its frequency.

*3.3 Scenarization by "recomposed documents"*

SCAC markup model allows us to obtain different fragments correlatives to concepts, they can therefore be transcluded (Nelson, 1999). In other words, it becomes possible to decontextualize them from their original article (the source document) to recontextualize them within a new document, a "recomposed document" with a singular meaning. This document recomposition is possible thanks to XSLT stylesheets, which operate both to filter fragments of information, sort, organize and arrange them according to the device designs. In this case and at present, the "hypermediator website" offers "concepts records" that assemble in a single interface all the fragments of information related to a concept, and categorize them according to the nature of the tags and therefore of their meaningless.

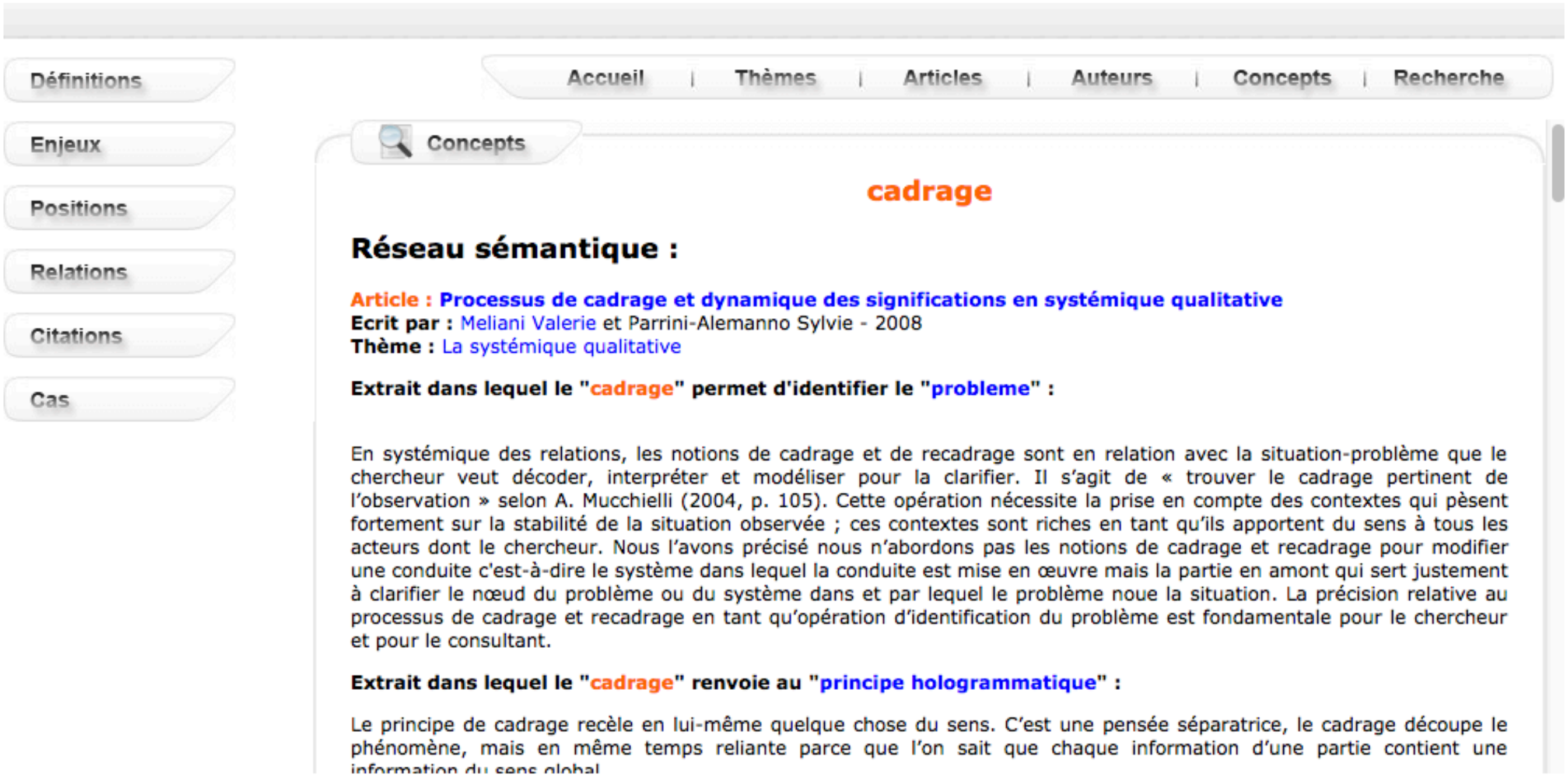

*Fig. 2. Example of "concept record". Here fragments of information outlining the relations between the concept of "framing" and other concepts.*

If in practice the principle of "decontextualization-recontextualization" is akin to making a copy / paste of fragments of information, it is distinguished by the traceability of information fragments to their original article. This traceability of information fragments respects the intellectual property of the authors but also allows the reader to consult the article source if necessary. The scenarization chosen in order to present fragments to the readers - the concepts records - goes also in this sense. Moreover, it has been developed so as to accompany the reader in the information hyperspace and in the building of his knowledge.

Although interesting at this stage of development, the hypermedaitor website remains too borrowing of the volume bookish and does not allow the reader to have a meta view of the conceptual univers of a knowledge domain.

## 4. The complex systems approach

### *4.1 Fundamental principles of the complex systemic : from human systems to knowledge systems*

The paradigm of complexity and the contributions of the systemic approach allows us to take a fresh look at knowledge systems in the direct lineage of ideas precursors of hypertext. The knowledge of a given field of knowledge are not sequential, although it is structured by authors around concepts in paragraphs, articles, they are an all connected: they are the institution of domain knowledge. And the holistic vision of a knowledge system and its constituent subsystems offered by the complex systems approach allows us to envisage a new angle on the organization of knowledge and its visualization.

The complex systemic vision is generally relevant in order to report and understand a complex reality co-constructed by actors in a situation of interaction. The researcher models the system and subsystems that are maintained to provide an understanding of the structure and dynamics of interactions that make the situation. From our point of view, this structural-functional conception of the systemic necessarily adorns the idea of collective construction of meaning. Therefore, the system is understood as an "*intelligible tangle and finalized of interdependent activities*" and can collect both a phenomenon in its consistency, in its unity and its internal interactions (Le Moigne, 1999, p 30), and can be described in terms of decomposable elements, and almost decomposable or undecomposable.

We assume the idea of decomposability in order to argue that by studying the meanings for this co-contruction of reality, systemic approach may also be relevant to represent and analyze the construction of meaning in a "relevant system" be it a communication system or a system knowledge.

In this sense, the systemic, as it applies to organize in understanding of communicative action of the human groups, can equally help organize and report the semantic construction of a knowledge global system. To do this we are building on one of the fundamental axioms raised by the School of Palo Alto (Watzlawick, Beavin & Jackson, 1979). In the acceptance of a complex systemic, communication consists of two dimensions: content (information) and the relationship (link). Thus, it is between these two dimensions that the meaning emerges by contextualizing informations and adding, by the relationship, a meta-communication. So, informations and their linkage, their relations, would allow to organize and to give sense with the aim of their understanding in the global system.

This complex vision includes in particular the hologrammatic principle that postulates that the systemic "everything" is included in the part and the part in the all (Morin & Le Moigne, 1999) making central the idea of "reliance" (Morin, 1990). Therefore, our thesis is the following: if the complex systemic allow to explore in understanding the meanings relating to a co-construction of "reality" situational and communicative, it may also be relevant to represent and analyze the semantics a field of knowledge. This, by interesting the "informations" and the "relations" between knowledge (micro-all) and how they are organized to construct a system of knowledge (all) relevant. The parties each have their singularity but are nonetheless "fragments" of the all, they are "micro-all" virtuals. The principle hologrammatic transposed to information returns to postulate that the knowledge is included in the understanding and the understanding in the knowledge. The knowledges each have their singularity but are also "particles" (concepts) of knowledge, they are "plots to know" that we need to organize and to connect in order to make sense.

*4.2 Systemic modelization for knowledge*

Whether to participate in a situation, to solve a problem, to understand a phenomenon, to understand a concept, or to build aknowledge, man organizes his thought according to schemes. These schemes are constructed on the basis of their recurrences and allow him to recognize, to understand and therefore to "know". According to Valéry, to understand a phenomenon, man can not escape develop a process "natural" modelization. Therefore, it seems quite plausible to transpose "artificially" this natural organization phenomenon.

The systemic approach induces this modelization process to "think the phenomenons using system", then it allows to consider to design, to visualize and to process knowledge organization. Its universal character allows from semantics of a domain of knowledge to modelize the socio-scientific thought and to reveal "patterns". The "patterns" (Simon, 1979, 1981) are models organizers, are recurring complex schemes (Bateson, 1979) used to represent an ontological thought.

Patterns (Simon, 1979, 1981) are organizing models, are recurring complex shapes (Bateson, 1979) used to represent an ontological thought. *"The pattern is an organized form and a structuring form identified by a cognitive act of perception"* (Le Moigne, 1999, p 47). So based on this ontological thought - the "pattern recognition" (Ibid.) became an anthropological culture and then an semantical culture organized and structured. It becomes possible to think of the complex construction and the complex restitution of systems knowledge on forms that Le Moigne calls "gestalt patterned" involving inseparability of product and process (Ibid.)

While it is undeniable that the process semantization are inherently highly complex, unpredictable and unstable, they are nevertheless modeled in a "common cultural contextual framework" if one refers to the definition of Le Moigne *"modelize is a share of elaboration and intentional construction by composition of symbols, models that make intelligible a complex phenomenon and amplify the reasoning of the actor throwing a deliberate action on the phenomenon; reasoning aiming to anticipate consequences of these possible actions projects"* (1999, p.5). This definition of the modelization process seems fruitful to think knowledge modelization in a given field. For the reader the field of knowledges is a complex system composed of texts, authors networks, schools of thought, concepts, contexts, on which involved the societal effects across spatial and temporal contexts. Here the hypermediator website offers models of knowledge organization precisely to amplify the reasoning of the reader, the knowledge acquisition, the assimilation of knowledges, understanding of phenomena or action.

*4.3 Conceive intellectual paths in a domain of knowledge*

According to Simon, our memory works on the accumulation of traces formed on our experiences, our past knowledges, that we will recombine in case of problem situation. This recombination of traces constitutes what he calls "pathing map". Each recombination, depending on the problems to solve and from the context generates a more appropriate different recombination based on these traces which "form" the basic structure of our knowledge. Basic structure according to which, by organizing the elements, we can build an infinity number of maps, each map is then seen as a *pattern* in which the problem becomes resolvable or intelligible. That is in seeking to uncover, to model, via the SCAC markup model these traces, that we aim through the hypermediator website and the markup work to offer "pathing maps" in the complex system of a knowledge field by respecting the "natural" process of discovery, of design and of construction of knowledge.

If the idea of "map" does not fail to resonate with the idea of visualization, Simon describes the memory as a library of symbols, of patterns, of shapes, of metaphors that can combine, recompose to infinity and inform reciprocally (Demailly, 1999). If the human memory is not "extensible" no one today denies that the techniques can play the role of external memory, even of "universally shared library" in reference to Otlet, library of signs, symbols, knowledge, no longer congealed in the shackles of volume but dynamic through networking and sharing, a form of collective memory in which it would be possible to propose courses of reading. The hypermadiator website, the semantic markup, can concretely conceive of this library, they can it offer directly as a map thanks to information visualization.

How to build maps of pathing? The SCAC markup model can point interconnections between concepts but mostly it focuses on the typification of semantic relations. This markup model offers a structure correlatively to the meaning units identified in the documents, the concepts' relations can provide a modelization, a mapping of knowledge domain. As for systemic study of human systems, that is recurring behavior, the relational form that repeats which is interesting. However to systemic practice, omit the recessive facts or isolated is a simplifying. But the use of statistical weighting based on strong ties between recurrences, moderate or low, as developed in the work of Simon (1977) will help reveal the complex architecture of knowledge.

## 5. Hypermediator website and infovizualisation

Following the precepts of the systemic approach, we sought to model the conceptual universe outlined in our corpus. The latter is composed of 33 scientific articles in the field of information and communication sciences, which represents more than 500 pages. These articles are on the theme of communication on organizations.

The data analysis from the SCAC markup model has allowed us to track down inside the corpus 149 concepts, 100 interrelations between concepts. On these 100 interrelations, we can count 29 meronym/holonym relations, 6 hypernym/hyponym relations and 7 analog relations. The other interrelations between concepts are associative ones. To view the system of knowledge extracted from our corpus, we used the free software Gephi (https://gephi.org/). Gephi allows to process data graphically and to analyzed it statistically. "*We are witnessing the emergence of a fourth paradigm of scientific research based on the intensive use of computation on large masses of numerical data*" (Bell, 2009, quoted by Le Crosnier, 2010, p.54). "*The data analysis covers a wide range of activities taking place at various stages of progress of these operations, including the use of databases (as opposed to sets of raw files on which the databases can access), analysis and modeling, and finally, the data visualization*" (Bell, 2009, quoted by Le Crosnier, 2010, p.54).

*5.1 Global system visualization*

The <position> and <relation> tags inherent in the SCAC markup model allow us to extract from the entire corpus of the relations between the concepts expressed by the authors of our corpus. Thus, we were able to modelize the global system with the concepts composing the corpus via statistical weighting of links (strong, moderate or weak) between nodes.

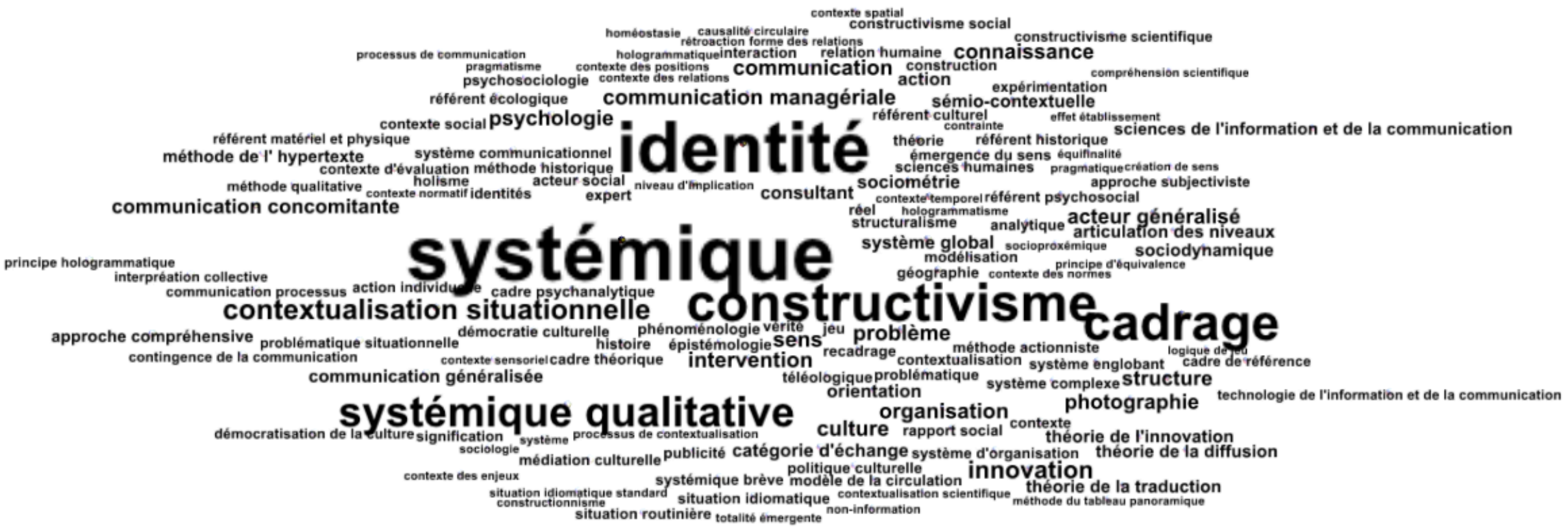

*Fig. 3. Visualization of the global system*

This statistical weighting relative to our sample only reveals four strong concepts whose recurrence is between 10 and 14, 3 concepts moderate (between 5 and 9 recurrences), the vast majority of concepts (142) with a low recurrence under 4.

*5.2 Focus on emergent patterns of the corpus*

The statistical weighting also allows us to make emerge the dominant patterns from our corpus. The figure 4 highlights the existing interconnections around a strong concept of "systemic". The links between concepts are more or less marked the one hand according to recurrence but also according to the consensus among the authors on the links typification.

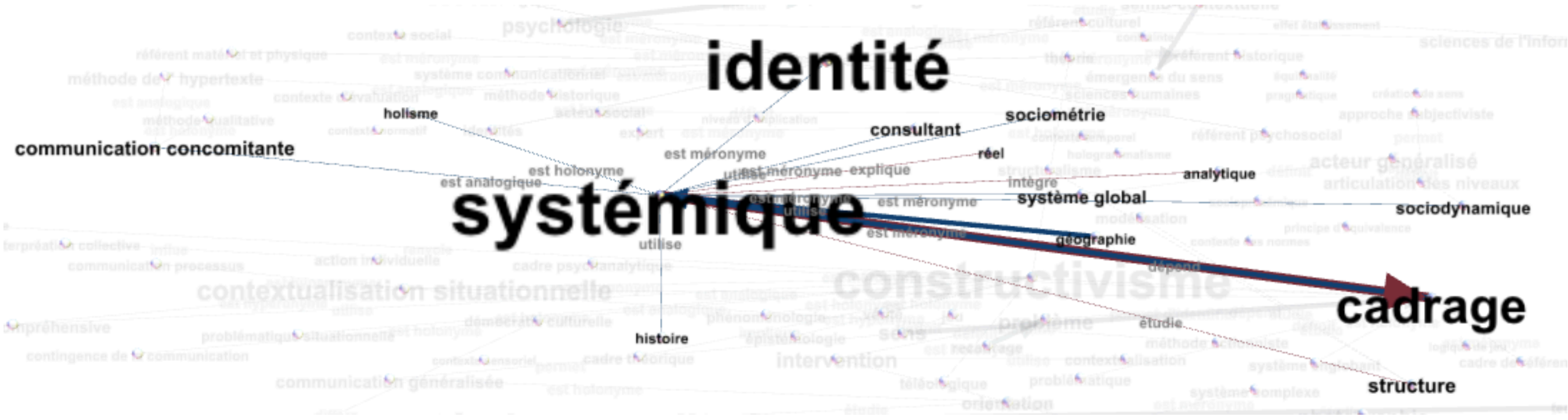

*Fig. 4. Pattern "Systemic"*

We note, in this regard, the consensus on the concept relations between "systemic" and "framing". For collective intelligence inherent in our corpus, the concept of "systemic" "depends" on the concept of "framing".

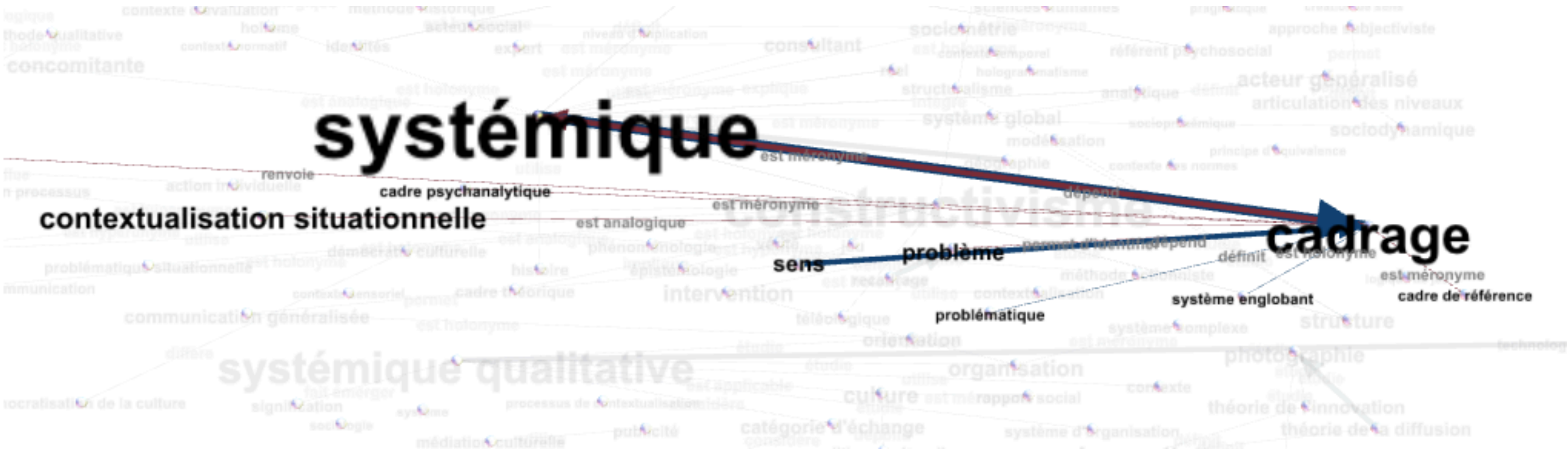

*Fig. 5. Pattern « framing »*

The study of the pattern "framing" allows to highlight the relation that the concept of "framing" maintains with concept of "problem": the "framing" "identifies" the "problem." According to this map pathing we understand that given the collective intelligence that the “systemic depends on framing, itself needed to identify the problem”.

*5.3 Perspectives of navigation for the reader*

Thus, the statistical weighting can not only reveal the complex architecture of knowledges of a domain, but also allows you to focus on the patterns that compose it. These patterns are all paths, possible pathways for the reader. The information visualization of the overall system and patterns allows the reader to navigate freely within the information space according to their interests while offering courses of reading constructed which were formalized using the collective intelligence. Thus, a hypermediator website offers the reader two ways to access information. The first is based on recomposed documents by the fragments of information through "concepts records", which are ultimately an aggregation of informational contents previously categorized and traceable. The second is based on an information visualization of a knowledge system, which allows the reader to understand the global or specific structure of a domain. The two modes of access to information that we have just mentioned are obviously interacting, the actor may at any time moved from the information visualization to the recomposed documents and vice versa. As such, the hypermediator website is fully involved in the co-contruction of knowledge to the reader and can claim the title of technology of intelligence.

## 6. Conclusion

Thus, our position is clear, we support the idea of a return to the sources, to the organization of the system of thought, "communication" of knowledges and ideas, and to the construction of meaning as an iterative process of knowledges construction. In addition, we consider that the organization of knowledges, the knowledges construction is part of a process of knowledge modeling. In doing so, we argue the information visualization as an "artificial" model close from intra-psychic processes that organize knowledges, that formalize the relations between knowledges proposed in a non-sequential manner but networking and taking into account linkages, relations between the units of meaning, to navigate through the content without losing the fundamental idea of constructed meaning.

Indeed, if the technique allows us to capitalize the informations, the process of knowledge construction, the organization of "natural knowledges" can not but take into account the complex processes of constructing meaning. The semantic is in the heart of the project, the hypermediator website, such as we organize it, joins the initial prospects of the hypertext and the semantic web. It has to be considered as a technology of intelligence for the benefit of knowledge. Therefore, thanks to the "semantic markup", the potentials of such a mediation space allow to conceptualize, to link, to organize, to capitalize, to restore, to enhance, to transmit a world of knowledge by taking into account, first of all, the human complex process on meaning construction.

Finally, if our current corpus is far from representing an important mass of data, its treatment nevertheless suggests interesting perspectives to understand and analyze the knowledges domain in question, or to interview several areas of knowledges in order to extract concepts acting as "boundary objects" (Carlile, 2002, 2004) between these areas.